\documentclass[11pt]{article}
\usepackage[preprint]{acl} % change to review for submission and final for camera-ready
\usepackage[utf8]{inputenc}
\usepackage[LGR,T1]{fontenc}
\usepackage{textalpha}
\usepackage{newtxtext,newtxmath}
\DeclareFontFamilySubstitution{LGR}{ntxtlf}{cmr}
\usepackage{microtype}
\usepackage{graphicx}
\usepackage{booktabs}
\usepackage{hyperref}

\title{Automatic Annotation of Ancient Greek Vowel Length}
\author{
  \textbf{Albin Thörn Cleland}\textsuperscript{1},
  \textbf{Eric Cullhed}\textsuperscript{2}
\\
\\
  \textsuperscript{1}Centre for Languages and Literature, Lund University, Sweden \\
  \textsuperscript{2}Department of Linguistics and Philology, Uppsala University, Sweden \\
  \small{
    \textbf{Correspondence:} \href{mailto:albin.thorn_cleland@klass.lu.se}{albin.thorn\_cleland@klass.lu.se}
  }
}

\newcommand{\ogamacronized}{\url{https://huggingface.co/datasets/Ericu950/oga-macronized}}

\newcommand{\ogamacronizerchar}{\url{https://huggingface.co/Ericu950/oga-macronizer-char}}

\newcommand{\norma}{\url{https://huggingface.co/datasets/Urdatorn/norma}}

\newcommand{\hypotacticcrawl}{\url{https://github.com/Urdatorn/hypotactic}}

\newcommand{\ogamacronizercharseedseventeen}{\url{https://huggingface.co/Ericu950/oga-macronizer-char-seed17}}

\newcommand{\hexameter}{\url{https://github.com/Urdatorn/scan-macronized}}

\newcommand{\baselinelongaccuracy}{0.0}
\newcommand{\baselineshortaccuracy}{100.0}
\newcommand{\baselinemixedaccuracy}{83.1}

\newcommand{\rulebasedlongaccuracy}{53.9}
\newcommand{\rulebasedshortaccuracy}{69.3}
\newcommand{\rulebasedmixedaccuracy}{66.7}

\newcommand{\transformerlongaccuracy}{64.7}
\newcommand{\transformershortaccuracy}{97.6}
\newcommand{\transformermixedaccuracy}{92.0}

\newcommand{\baselinelongprecision}{0.0}
\newcommand{\baselineshortprecision}{83.1}
\newcommand{\baselinemixedprecision}{83.1}

\newcommand{\rulebasedlongprecision}{89.2}
\newcommand{\rulebasedshortprecision}{97.1}
\newcommand{\rulebasedmixedprecision}{96.0}

\newcommand{\transformerlongprecision}{85.0}
\newcommand{\transformershortprecision}{93.4}
\newcommand{\transformermixedprecision}{92.3}

\newcommand{\baselinelongrecall}{0.0}
\newcommand{\baselineshortrecall}{100.0}
\newcommand{\baselinemixedrecall}{83.1}

\newcommand{\rulebasedlongrecall}{53.9}
\newcommand{\rulebasedshortrecall}{69.3}
\newcommand{\rulebasedmixedrecall}{66.7}

\newcommand{\transformerlongrecall}{64.7}
\newcommand{\transformershortrecall}{97.6}
\newcommand{\transformermixedrecall}{92.0}

\newcommand{\baselinelongfone}{0.0}
\newcommand{\baselineshortfone}{90.8}
\newcommand{\baselinemixedfone}{83.1}

\newcommand{\rulebasedlongfone}{67.2}
\newcommand{\rulebasedshortfone}{80.9}
\newcommand{\rulebasedmixedfone}{78.7}

\newcommand{\transformerlongfone}{73.5}
\newcommand{\transformershortfone}{95.4}
\newcommand{\transformermixedfone}{92.2}

\newcommand{\baselinelongcoverage}{100.0}
\newcommand{\baselineshortcoverage}{100.0}
\newcommand{\baselinemixedcoverage}{100.0}

\newcommand{\rulebasedlongcoverage}{64.1}
\newcommand{\rulebasedshortcoverage}{70.6}
\newcommand{\rulebasedmixedcoverage}{69.5}

\newcommand{\transformerlongcoverage}{98.8}
\newcommand{\transformershortcoverage}{99.9}
\newcommand{\transformermixedcoverage}{99.7}

\begin{document}
\maketitle
\begin{abstract}
  Prior work in Ancient Greek NLP relies on corpora that do not disambiguate the phonemic vowel length of α (alpha), ι (iota), and υ (ypsilon), together known as the dichrona. Depending on lexeme, morphology, sandhi, syntax, and conventions of period, genre, and verse form, each of these letters can represent either a long or a short vowel. Deciding and marking the correct length is known as ‘macronizing’, a long-tail problem given the sheer mass of word forms and the context dependency of individual instances. No macronized corpus of Ancient Greek is publicly available at scale, so a stand-alone macronizer is needed. While previous work has shown how to build a static, corpus-bespoke vowel-length dictionary, the present paper constructs the first general-purpose macronizer for arbitrary Ancient Greek input. Given input carrying lemma, part-of-speech, and morphological annotation in the standard CoNLL-U format, a set of recursive modules lets less common word forms inherit markup from more common forms of the same lexical word. The macronizer's chief application is generating training data for machine learning: we show that a small character-level transformer trained on the macronizer's own output learns to generalize past the cases the rule-based system leaves unmarked, matching or exceeding its accuracy on a gold-standard, manually annotated benchmark of verse and prose. We also show that macronization can improve downstream prosodical NLP tasks like verse scansion. 
\end{abstract}

\section{Introduction}
\subsection{Vowel lengths and the dichrona}
Phonemic vowel length divides a language's vowels into two contrasting classes whose distinction bears on sense \citep{oddenRepresentationVowelLength2011}. Ancient Greek belongs to the type of language in which the length distinction is necessarily accompanied by a distinction in vowel quality.\footnote{The exact nature of the contrast depends on the way the members of the two length classes are mapped to each other, which can be of three general kinds. In languages like Finnish, Thai, and Japanese, long and short vowels have the same quality. In other languages, like Swedish and Ancient Greek, length distinction is (or was) necessarily accompanied by quality distinction \citep[97]{probertPhonology2010}. Others still, like Cantonese, have a mapping which is nontrivial, and analyses may disagree as to what are taken as the contrasting pairs \citep{luoPerceptionCantoneseVowel2019}.}

The standard Ancient Greek script does not disambiguate the phonemic vowel length of all vowel pairs, leaving α (alpha), ι (iota), and υ (ypsilon), together known as the \emph{dichrona} (of two quantities), with just one letter per pair.\footnote{Alphas, iotas, and ypsilons appearing as part of diphthongs or bearing the circumflex accent are not considered dichronic, as their length is always long in that context.} The three dichrona can, depending on lemma, morphology, sandhi, syntax, and conventions relative to dialect, period, genre, and metre, represent either a long or a short vowel. For example:\footnote{The division is an artificial simplification: any given case will depend on all of these factors.}

\begin{itemize}
  \item Lexeme: the iota in ἱμάτιον is long while the iota in φίλος is short.
  \item Morphology: the ypsilon in the nominative κῆρυξ is short, but the ypsilon in the genitive κήρυκος is long.
  \item Sandhi: the first alpha in τἄλλα is long, whereas without crasis τὰ ἄλλα has no long alpha.
        % TODO(Albin): ἄν's definition is stated inconsistently across this paper
        % too (cf. S2's "modal particle"/subjunctive framing). See the fuller note
        % on this in macronizer_arxiv.tex -- same fix needed here.
  \item Syntax: the alpha in ἄν is long whenever appearing in a conditional clause without εἰ, but short everywhere else.
  \item Dialect: the iota in ὄρνις can be short in many dialects, but in fifth-century spoken Attic it was arguably consistently long.\footnote{E.g.\ Aristophanes, \emph{Birds} 103; see \citet{pizzoneOccasionalityByzantineDidacticism2024} for later Greek.}
  \item Period: σκνιπός has long iota in the sixth century B.C. iambic trimeter of Semonides, but short in the trimeter of Gregory of Nazianzus, written 900 years later.\footnote{Semonides in \emph{Etymologicum Magnum} 270, 45; Gregory of Nazianzus, \emph{Εἰς τὰ ἔμμετρα} 7.}
  \item Genre: the iota in τίνω is long in epic and short in drama and prose.
  \item Metre: in the hexameter line καλὰ μὲν ἠέξευ, καλὰ δ᾽ ἔτραφες, οὐράνιε Ζεῦ, (Callimachus, \emph{Hymn to Zeus} 55) the root alpha of καλά is long in princeps position and short in biceps.
\end{itemize}
The process of determining and marking these lengths is known as macronizing. In verse, macronization is a prerequisite for metrical analysis: Greek poetry regulates syllable weight, which depends primarily on vowel length. In prose, it disambiguates orthographic minimal pairs such as ἄν, which if long introduces an iterative temporal clause but if short is the modal particle, or ἵσταμεν, imperfect with long (augmented) iota but otherwise present -- distinctions any syntactic or morphological analysis needs to get right.

\subsection{Research Questions and Scope}
\begin{itemize}
  \item RQ1: How much of a large unseen Ancient Greek corpus can be macronized by algorithms and databases alone, given that the input is first marked up with lemma and morphology?
  \item RQ2: Can the coverage and quality of macronization be improved further by training a neural token classifier on an algorithmically macronized corpus?
  \item RQ3: Does macronization improve performance on other tasks important to Ancient Greek NLP, such as automatic scansion?
\end{itemize}
\subsection{Earlier Ancient Greek NLP and Vowel Length}
We are not aware of any earlier Ancient Greek machine learning project, including \citet{sansomSEDESMetricalPosition2023}, \citet{kardosOdyCy2023}, and \citet{celanoStateoftheArtMorphosyntacticParser2024}, having tried to macronize its corpus. \citet{wingeAutomaticAnnotationLatin2015} pioneered vowel-length annotation in classical languages with a static macronizer for Latin that agrees with human annotators 98\% of the time; \citet{richterAssessedAnnotatedVowel2025} exemplify recent progress on a modern language with a similar two-way length contrast. \citet{thornclelandMacronizerAncientGreek2024} showed how to algorithmically build a static vowel-length dictionary for Greek, but was limited to a small set of texts (the extant tragedies) and leveraged only a few of the possible sources of disambiguation. We know of no previous attempt to use machine learning for the actual classification of vowel lengths in classical languages.

\section{A Rule-Based Macronizer}
The goal is for every long dichronon in the input to be followed by an underscore ("\_") and every short dichronon by a caret ("\textasciicircum{}"); dichrona the macronizer fails to disambiguate are left unmarked. As the target corpus to optimize for, we use the OGA (Opera Graeca Adnotata),\footnote{\url{https://github.com/OperaGraecaAdnotata/OGA}} a combination of First1KGreek, PatristicTextArchive, and Perseus' canonical-greekLit. Development is corpus-driven: each prototype is run over the OGA, and the resulting logs and statistics guide the tuning of the next version.

Word forms attested with identical morphological tags but different lengths (ὄρνις, σκνιπός, τίνω above) are assigned the ancient Attic length across the board, leaving exceptions to downstream tools such as metrical analysis. For forms like ἀεί, whose length divides across no obvious genre or period fault line, we exploit the fact that around seven out of ten dichrona in the Attic classics are short: defaulting such forms to short performs better on average.

Input is first marked up with lemma, part-of-speech, and morphological tags by odyCy \citep{kardosOdyCy2023}. A handful of forms odyCy mishandles (e.g., the crasis τἄλλα) are macronized preemptively, and ἄν receives dedicated logic: long ἄν, being a crasis of ἐάν, requires a subjunctive verb and the absence of εἰ in its clause, which together yield a necessary and sufficient condition. All remaining ambiguous tokens then flow through thirteen macronization modules of three types, applied in a fixed order of dependability so that later modules never overwrite earlier ones.

\paragraph{Database modules.} A dictionary of 424,823 distinct ambiguous word forms sieved from the conjugation and declension tables of 40,000 Wiktionary pages; 34,229 lemmata (entry forms only) macronized from Liddell--Scott--Jones; 60,660 lines of scanned verse from Hypotactic \citep{chamberlainHypotacticcom2023} (given lowest priority since the non-standard division of consonants between codas and onsets has been algorithmically standardized by us and is not perfect); and a hand-curated custom database of 567 common word forms not covered elsewhere.

\paragraph{Algorithmic modules.} The accentuation laws of classical Attic (the two halves of the σωτῆρᾰ-rule and the short ultima of proparoxytones), which constrain vowel length practically without exception; six general rules for nominal and verbal endings distilled from standard declension tables (e.g., first-declension feminine accusative singular -αν is long, dative -ι is short); and prefix analysis: if a lemma minus a known prefix is still an LSJ entry, the prefix is macronized, so ἀφίκοντο gets a short alpha because ἰκνέομαι is in LSJ.

\paragraph{Recursive modules.} When a token fails lookup, recursion retries normalized variants: recovering oblique cases from attested nominatives (στρατηγόν is macronized even if only στρατηγός is in the databases), removing second accents from enclitic-bearing words like Καλλίμαχός, undoing elision, converting grave to acute, and decapitalizing with custom casing functions.

\section{A Transformer Macronizer}

\paragraph{Silver training data.}
The macronizer's chief application is generating training data. We ran it over the forty-million-token OGA corpus (1,999 works), converting OGA's shipped CoNLL-U lemma, POS, and morphology annotation \citep{celanoStateoftheArtMorphosyntacticParser2024} directly into the macronizer's internal representation rather than re-running odyCy inference. The result is 2,234,599 macronized sentences with 69.20\% of open-syllable dichrona resolved (11,627,591 of 16,802,080, answering RQ1), published as a HuggingFace dataset.\footnote{\ogamacronized}

\paragraph{Architecture.}
Vowel length is an intrinsically character-level property, and character-level transformers have already proven apt for Ancient Greek in epigraphic restoration \citep{assaelRestoringAttributingAncient2022}. We therefore train a small character-level encoder of roughly 0.87M parameters (four layers, hidden size 128, four heads, intermediate size 512), released on the Huggingface Hub.\footnote{\ogamacronizerchar} Each character is factored into two planes with separate learned embeddings summed with a positional embedding: a \emph{letter} plane (the 24 Greek letters with sigma variants folded, space, and a catch-all "other" for punctuation, digits, digamma, and non-Greek characters) and a \emph{diacritic} plane whose small vocabulary of accent/breathing/diaeresis combinations is fitted from data. A three-way head predicts none/short/long at genuinely ambiguous dichrona; during training the diacritic plane is randomly stripped for a fraction of characters, so the model has also had to macronize unaccented input.

\paragraph{Training on silver labels.}
The most consequential design choice concerns the positions that the rule-based system leaves unmarked: wherever the rule-based macronizer commits to a vowel length, that judgment becomes a training label; wherever it abstains, the position is excluded from the loss by ignore-index masking, and never treated as an implicit "short." We trained on the full corpus with AdamW and a one-cycle schedule on a single A100, holding out 10\% of lines for validation and early stopping (patience of eight checkpoints).

\section{Results}

\paragraph{Benchmarking.}
To evaluate macronization quality in a way that counts not only coverage, but also right and wrong lengths (both the ``coverage and quality'' of RQ2), we introduce \emph{Norma Syllabarum Graecarum}, a manually annotated benchmark marking syllable boundaries, syllable weight, and vowel length in 1,228 lines of verse or prose from fourteen authors spanning the archaic to the imperial age, plus 299 lines from the responding lyric songs of Aristophanes taken from \citet{thornclelandHiddenChoralStimuli2025}. All the texts were chosen so as not to coincide with those included in Hypotactic as of 2025.\footnote{\norma. Authors: Alcman, Bacchylides, Aeschylus, Sophocles, Euripides, Aristophanes, Thucydides, Plato, Dioscorides, Plutarch, Epictetus, Origen, Nonnus, Quintus. See \hypotacticcrawl{} for the crawl of Hypotactic used.} Every open syllable containing a dichronon (α, ι, υ) is marked short or long; the benchmark tests open syllables only.

As a baseline to compare our two models against, we use a constant prediction of short. \autoref{tab:transformer-results} summarizes the results of evaluating both models and the baseline with the same benchmark and scoring methodology.

\begin{table*}[t]
  \centering
  \small
  \begin{tabular}{llrrrrr}
      \toprule
      System         & Label & Acc. (\%)                          & P (\%)                            & R (\%)                           & F1 (\%)                        & Cov. (\%) \\
      \midrule
      Constant-short & Long  & (\baselinelongaccuracy)            & (\baselinelongprecision)          & (\baselinelongrecall)            & (\baselinelongfone)            & (\baselinelongcoverage) \\
                     & Short & (\baselineshortaccuracy)           & \baselineshortprecision           & (\baselineshortrecall)           & \baselineshortfone             & (\baselineshortcoverage) \\
                     & Mixed & \baselinemixedaccuracy             & \baselinemixedprecision           & \baselinemixedrecall             & \baselinemixedfone             & (\baselinemixedcoverage) \\
      \midrule
      Rule-based     & Long  & \rulebasedlongaccuracy             & \textbf{\rulebasedlongprecision}  & \rulebasedlongrecall             & \rulebasedlongfone             & \rulebasedlongcoverage \\
                     & Short & \rulebasedshortaccuracy            & \textbf{\rulebasedshortprecision} & \rulebasedshortrecall            & \rulebasedshortfone            & \rulebasedshortcoverage \\
                     & Mixed & \rulebasedmixedaccuracy            & \textbf{\rulebasedmixedprecision} & \rulebasedmixedrecall            & \rulebasedmixedfone            & \rulebasedmixedcoverage \\
      \midrule
      Transformer    & Long  & \textbf{\transformerlongaccuracy}  & \transformerlongprecision         & \textbf{\transformerlongrecall}  & \textbf{\transformerlongfone}  & \textbf{\transformerlongcoverage} \\
                     & Short & \textbf{\transformershortaccuracy} & \transformershortprecision        & \textbf{\transformershortrecall} & \textbf{\transformershortfone} & \textbf{\transformershortcoverage} \\
                     & Mixed & \textbf{\transformermixedaccuracy} & \transformermixedprecision        & \textbf{\transformermixedrecall} & \textbf{\transformermixedfone} & \textbf{\transformermixedcoverage} \\
      \bottomrule
  \end{tabular}
  \caption{Results on the Norma Syllabarum Graecarum benchmark (1,916 gold positions), reported separately for long, short, and mixed labels. Acc.: accuracy; P: precision; R: recall; Cov.: coverage. For the constant baseline, trivial numbers have been put inside parentheses. The best non-trivial result for each metric and label is in bold face.}
  \label{tab:transformer-results}
\end{table*}

That the rule-based macronizer has the highest precision confirms that it works as intended; namely as a reliable but conservative reflection of general rules and written sources. Likewise, the transformer's ability to generalize is shown by the increase in coverage from 69.5\% to 99.7\% while dropping less than five percent in precision for each label and increasing in recall almost ten percent for long and more than twenty-five percent for short. The tranformer's genuine improvement over the rule-based model is best shown by the increase in F1 of 6.3 points for the difficult long label, and is further borne out by the fact that only the transformer improves upon the baseline's mixed F1 (by 9.1 points), notwithstanding the very uneven distribution of labels.\footnote{An additional transformer trained with a different seed gets the exact same scores as in \autoref{tab:transformer-results} (available at \ogamacronizercharseedseventeen).} Together, these results answer RQ2 in the affirmative: a transformer classifier trained on an algorithmic macronizer's output can generalize past the cases the rule-based system leaves unmarked, matching or exceeding its accuracy on a gold-standard benchmark.

\paragraph{Downstream task: scansion.}
A common task in Ancient Greek NLP is scansion of quantitative poetry, which, at minimum, means predicting for each input verse line a metrical pattern of heavy and light syllables, such as the following example of dactylic hexameter:

\begin{quotation}
  \texttt{-uu-uu-uu-uu-uu--}.\footnote{A finer analysis could also predict the word boundaries, making visible the caesurae.}
\end{quotation}
To investigate the effects of macronization, we fine-tuned two identical distilbert-based \citep{sanh2019DistilBERTAD} classifiers on 41,695 lines of hexameter from Hypotactic. The classifiers were trained to classify each line as one of the thirty-one metrical hexameter patterns present in the corpus, with one model getting input lines macronized by our transformer and the other not, and both using the same 80/10/10 split stratified by source texts.

Over three different seeds, macronization improved the accuracy of the scansion classifier by a mean $5.8\% \pm 1.5\%$ (95\% t-interval).\footnote{Each model trained for thirty epochs, and the best validation performance peaked between epochs twenty-four and twenty-eight. The scansion classifiers, data, and recipes used for training them are all available at \hexameter.} This result answers RQ3 in the affirmative: macronization improves downstream scansion accuracy.

\section{Summary}

Returning to the research questions posed at the outset: first, the rule-based pipeline alone disambiguates 69.2\% of open-syllable dichrona across the forty-million-token OGA corpus and reaches \rulebasedmixedaccuracy\% mixed-label accuracy and \rulebasedmixedfone\% mixed-label F1 accuracy on Norma. Second, a character-level transformer of under a million parameters, trained purely on this silver output with abstentions masked from the loss, reaches \transformermixedaccuracy\% mixed-label accuracy and \transformermixedfone\% mixed-label F1 on the same benchmark while covering 99.7\% of positions. Third, macronization improves downstream scansion accuracy by around five points. We hope that other applications, such as stylometry and morphology tagging, can benefit as we release the macronized corpus, both models, and the benchmark open-source as a foundation for further work on Ancient Greek NLP.

\section*{Limitations}

\paragraph{Attic normalization.} Forms attested with identical
morphology but divergent lengths are assigned the ancient Attic value
throughout. This is a deliberate simplification, and it is wrong by
design for Homeric, Doric, Hellenistic, and Byzantine text, and for
forms whose length divides along genre lines (e.g.\ τίνω). The
macronizer should at present be understood as an Attic-normalizing
system, and its output on non-Attic corpora carries a systematic bias
we have not quantified. This could be mitigated by taking account of metadata.

\paragraph{Transformer hyperparameters.} No hyperparameter
search was performed for the transformer macronizer.

\paragraph{Double dependence on upstream annotation.} The rule-based pipeline consumed
lemma, POS, and morphology from two external taggers: a fine-tuned Trankit \citep{celanoStateoftheArtMorphosyntacticParser2024} to make training data from OGA, and odyCy \citep{kardosOdyCy2023} for benchmarking on Norma Syllabarum Graecarum.\footnote{The Trankit used for OGA reports benchmarking POS 96.41, LAS 77.10, and LEMMA 91.41 \citep[52f]{celanoOperaGraecaAdnotata2026}, odyCy reports POS 95.39, LAS 73.09 and LEMMA 83.20 \citep[131]{kardosOdyCy2023} on the test fold of UD Proiel.} Errors in the
annotation propagate directly into macronization, and we have not
measured how much of the residual error is attributable to Trankit's
mistakes rather than to the rule-based macronization modules themselves. Analogously, the transformer macronizer is
trained exclusively on its rule-based teacher's judgments, so wherever the
rule-based system is systematically rather than randomly wrong (e.g.\ a
mistaken declension rule or an erroneous Wiktionary entry) the student
has no signal from which to recover. Its errors are correlated with
its teacher's, and the ceiling it approaches is the teacher's ceiling
rather than the gold standard's. Due to the long-tailedness of vowel lenghts, solving this problem would require fine-tuning on a considerable amount of manually corrected data.

% Add back for accepted/preprint version

\section*{Acknowledgments}
Computational resources were provided by the National Academic Infrastructure for Supercomputing in Sweden (NAISS), funded by the Swedish Research Council.

\bibliography{macronizer_arxiv}

\begin{thebibliography}{15}
\providecommand{\natexlab}[1]{#1}

\bibitem[{Assael et~al.(2022)Assael, Sommerschield, Shillingford, Bordbar,
  Pavlopoulos, Chatzipanagiotou, Androutsopoulos, Prag, and
  de~Freitas}]{assaelRestoringAttributingAncient2022}
Yannis Assael, Thea Sommerschield, Brendan Shillingford, Mahyar Bordbar, John
  Pavlopoulos, Marita Chatzipanagiotou, Ion Androutsopoulos, Jonathan Prag, and
  Nando de~Freitas. 2022.
\newblock \href {https://doi.org/10.1038/s41586-022-04448-z} {Restoring and
  attributing ancient texts using deep neural networks}.
\newblock \emph{Nature}, 603(7900):280--283.

\bibitem[{Celano(2024)}]{celanoStateoftheArtMorphosyntacticParser2024}
Giuseppe G.~A. Celano. 2024.
\newblock \href {https://doi.org/10.48550/arXiv.2410.12055} {A state-of-the-art
  morphosyntactic parser and lemmatizer for {Ancient Greek}}.

\bibitem[{Celano(2026)}]{celanoOperaGraecaAdnotata2026}
Giuseppe G.~A. Celano. 2026.
\newblock \href {https://doi.org/10.11588/DCO.2026.12.112290} {{Opera Graeca
  Adnotata}: Building a 40m+ token multilayer corpus for {Ancient Greek}}.
\newblock Bd. 12:46--60 Seiten.

\bibitem[{Chamberlain(2023)}]{chamberlainHypotacticcom2023}
David Chamberlain. 2023.
\newblock \href {https://hypotactic.com/} {hypotactic.com}.

\bibitem[{Kardos and Kostkan(2023)}]{kardosOdyCy2023}
Marton Kardos and Jan Kostkan. 2023.
\newblock \href {https://centre-for-humanities-computing.github.io/odyCy}
  {{odyCy}}.

\bibitem[{Luo et~al.(2019)Luo, Li, and Mok}]{luoPerceptionCantoneseVowel2019}
Jingxin Luo, Vivian~Guo Li, and Peggy Pik~Ki Mok. 2019.
\newblock \href {https://doi.org/10.1177/0023830919879471} {The perception of
  cantonese vowel length contrast by mandarin speakers}.
\newblock \emph{Language and Speech}, 63(3):635--659.

\bibitem[{Odden(2011)}]{oddenRepresentationVowelLength2011}
David Odden. 2011.
\newblock \href {https://doi.org/10.1002/9781444335262.wbctp0020} {The
  representation of vowel length}.
\newblock In Marc van Oostendorp, Colin~J. Ewen, Elizabeth Hume, and Keren
  Rice, editors, \emph{The Blackwell Companion to Phonology}. Wiley-Blackwell.

\bibitem[{Pizzone(2024)}]{pizzoneOccasionalityByzantineDidacticism2024}
Aglae Pizzone. 2024.
\newblock \href {https://doi.org/10.54103/interfaces-11-04} {The occasionality
  of {Byzantine} didacticism: a case study from the twelfth century ({Milan},
  {Veneranda Biblioteca Ambrosiana}, c 222 inf. f. 218r)}.
\newblock \emph{Interfaces: A Journal of Medieval European Literatures},
  11:51--73.

\bibitem[{Probert(2010)}]{probertPhonology2010}
Philomen Probert. 2010.
\newblock Phonology.
\newblock In Egbert~J. Bakker, editor, \emph{A Companion to the Greek
  Language}, pages 85--103. Wiley-Blackwell.

\bibitem[{Richter et~al.(2025)Richter, Fri{\dh}riksd{\'o}ttir, Bergsson, Maher,
  Benediktsd{\'o}ttir, and Gudnason}]{richterAssessedAnnotatedVowel2025}
Caitlin~Laura Richter, Kolbr{\'u}n Fri{\dh}riksd{\'o}ttir, Korm{\'a}kur~Logi
  Bergsson, Erik~Anders Maher, Ragnhei{\dh}ur~Mar{\'i}a Benediktsd{\'o}ttir,
  and Jon Gudnason. 2025.
\newblock \href {https://aclanthology.org/2025.nodalida-1.56/} {Assessed and
  annotated vowel lengths in spoken {Icelandic} sentences for {L1} and {L2}
  speakers: A resource for pronunciation training}.
\newblock In \emph{Proceedings of the Joint 25th Nordic Conference on
  Computational Linguistics and 11th Baltic Conference on Human Language
  Technologies (NoDaLiDa/Baltic-HLT 2025)}, pages 518--524, Tallinn, Estonia.
  University of Tartu Library.

\bibitem[{Sanh et~al.(2019)Sanh, Debut, Chaumond, and
  Wolf}]{sanh2019DistilBERTAD}
Victor Sanh, Lysandre Debut, Julien Chaumond, and Thomas Wolf. 2019.
\newblock Distilbert, a distilled version of bert: smaller, faster, cheaper and
  lighter.
\newblock \emph{ArXiv}, abs/1910.01108.

\bibitem[{Sansom and Fifield(2023)}]{sansomSEDESMetricalPosition2023}
Stephen Sansom, A. and David Fifield. 2023.
\newblock {SEDES}: Metrical position in {Greek} hexameter.
\newblock \emph{Digital Humanities Quarterly}, 17(2).

\bibitem[{Thörn~Cleland(2024)}]{thornclelandMacronizerAncientGreek2024}
Albin Thörn~Cleland. 2024.
\newblock A macronizer for {Ancient Greek}.
\newblock Master's thesis, Uppsala universitet.

\bibitem[{Thörn~Cleland(2025)}]{thornclelandHiddenChoralStimuli2025}
Albin Thörn~Cleland. 2025.
\newblock \href {https://doi.org/10.5281/zenodo.16603568} {Hidden choral
  stimuli: The role of accent in the refrains of {Aristophanes}}.

\bibitem[{Winge(2015)}]{wingeAutomaticAnnotationLatin2015}
Johan Winge. 2015.
\newblock Automatic annotation of {Latin} vowel length.
\newblock Master's thesis, Uppsala University, Uppsala.

\end{thebibliography}

\end{document}